\documentclass[article]{jss}
\usepackage{amsmath, amssymb, amsfonts} % for math symbols like \Sigma
\usepackage{algorithm}
\usepackage{algorithmic}
\usepackage{enumitem}                  % for removing space in itemize
\usepackage{graphicx, subfigure}  % for figures
\usepackage[capitalize]{cleveref}  % to cite mutiple fighres
\usepackage[utf8]{inputenc}          % for quotation

\newcommand{\Yb}{\mathbf{Y}} 
 
\newcommand{\Zb}{\mathbf{Z}}

\title{ \pkg{BDgraph}: An \proglang{R} Package for Bayesian Structure Learning in Graphical Models }
\author{ Reza Mohammadi \\ University of Amsterdam \And Ernst C. Wit \\ Universita della Svizzera Italiana }

\Plainauthor{ Ernst C. Wit, Second Author }                                 %% comma-separated
\Plaintitle{ BDgraph: An R Package for Bayesian Structure Learning in Graphical Models }           %% without formatting
\Shorttitle{ \pkg{BDgraph}: An \proglang{R} Package for Bayesian Structure Learning in Graphical Models } %% a short title (if necessary)

\Abstract{
Graphical models provide powerful tools to uncover complicated patterns in multivariate data and are commonly used in Bayesian statistics and machine learning. In this paper, we introduce an \proglang{R} package \pkg{BDgraph} which performs Bayesian structure learning for general undirected graphical models (decomposable and non-decomposable) with continuous, discrete, and mixed variables. The package efficiently implements recent improvements in the Bayesian literature, including that of \cite{mohammadi2015bayesianStructure} and \citet{dobra2018}. To speed up computations, the computationally intensive tasks have been implemented in \proglang{C++} and interfaced with \proglang{R}, and the package has parallel computing capabilities. In addition, the package contains several functions for simulation and visualization, as well as several multivariate datasets taken from the literature and are used to describe the package capabilities. The paper includes a brief overview of the statistical methods which have been implemented in the package. The main body of the paper explains how to use the package. Furthermore, we illustrate the package's functionality in both real and artificial examples. %, as well as in an extensive simulation study. 
}
\Plainkeywords{Bayesian structure learning, Gaussian graphical models, Gaussian copula, Covariance selection, Birth-death process, Markov chain Monte Carlo, G-Wishart, BDgraph, R} 
\Address{
  Reza Mohammadi, \\ Operation Management Section, \\ Faculty of Economics end Business, \\ University of Amsterdam, \\ Amsterdam, Netherlands, \\
  E-mail: \email{a.mohammadi@uva.nl} \\ URL: \url{ http://www.uva.nl/profile/a.mohammadi } 
 
  Ernst C. Wit, \\ Institute of Computational Science, \\ Universita della Svizzera Italiana, \\ Lugano, Switzerland, \\
  E-mail: \email{e.c.wit@rug.nl} \\ URL: \url{  http://www.math.rug.nl/~ernst/ }
}
\begin{document}
% - - - - - - - - - - - - - - - - - - - - - - - - - - - - - - - - - - - - - - - - - - - - - - - - - - - - - - - - - - - - - - - - - - - - - - - - - - - - - - - - - - - - |
\section{Introduction}
\label{sec:intro}

Graphical models \citep{lauritzen1996graphical} are commonly used, particularly in Bayesian statistics and machine learning, to describe the conditional independence relationships among variables in multivariate data. In graphical models, each random variable is associated with a node in a graph and links represent conditional dependency between variables, whereas the absence of a link implies that the variables are independent conditional on the rest of the variables (the pairwise Markov property). 

In recent years, significant progress has been made in designing efficient algorithms to discover graph structures from multivariate data \citep{dobra2011bayesian, dobra2011copula, jones2005experiments, dobra2018, mohammadi2015bayesianStructure, mohammadi2016bayesian, friedman2008sparse, meinshausen2006high, murray2004bayesian, pensar2017marginal, rolfs2012iterative, wit2015factorial, wit2015inferring, dyrba2018comparison, behrouzi2019detecting}. Bayesian approaches provide a principled alternative to various penalized approaches. 

In this paper, we describe the \pkg{BDgraph} package \citep{BDgraph} in \proglang{R} \citep{Rcoreteam} for Bayesian structure learning in undirected graphical models. The package can deal with Gaussian, non-Gaussian, discrete and mixed datasets. The package includes various functional modules, including data generation for simulation, several search algorithms, graph estimation routines, a convergence check and a visualization tool; see Figure \ref{fig:modules}. Our package efficiently implements recent improvements in the Bayesian literature, including those of \cite{mohammadi2015bayesianStructure, mohammadi2016bayesian, dobra2018, lenkoski2013direct, mohammadi2017ratio, dobra2011copula, hoff2007extending}. For a Bayesian framework of Gaussian graphical models, we implement the method developed by \cite{mohammadi2015bayesianStructure} and for Gaussian copula graphical models we use the method described by \cite{mohammadi2016bayesian} and \cite{dobra2011copula}. To make our Bayesian methods computationally feasible for moderately high-dimensional data, we efficiently implement the \pkg{BDgraph} package in \proglang{C++} linked to \proglang{R}. To make the package easy to use, the \pkg{BDgraph} package uses several \code{S3} classes as return values of its functions. The package is available under the general public license (GPL $\geq 3$) from the Comprehensive \proglang{R} Archive Network (CRAN) at \url{http://cran.r-project.org/packages=BDgraph}. 

In the Bayesian literature, the \pkg{BDgraph} is one of the few \proglang{R} packages which is available online for Gaussian graphical models and Gaussian copula graphical models. Another \proglang{R} package is \pkg{ssgraph} \citep{ssgraph} which is based on spike-and-slab proir.    On the other hand, more packages seem to be available in the frequentist literature. The existing packages include \pkg{huge} \citep{huge}, \pkg{glasso} \citep{glasso}, \pkg{bnlearn} \citep{scutari2009learning}, \pkg{pcalg} \citep{kalisch2012causal}, \pkg{netgwas} \citep{behrouzi2017netgwas}, and \pkg{QUIC} \citep{quic, hsieh2014quic}. 

In Section \ref{sec:get started} we illustrate the user interface of the \pkg{BDgraph} package. In Section \ref{sec:method} we explain some methodological background of the package. In this regard, in Section \ref{subsec:GGMs} we briefly explain the Bayesian framework for Gaussian graphical models for continuous data. In Section \ref{subsec:GCGMs} we briefly describe the Bayesian framework in the Gaussian copula graphical models for data that do not follow the Gaussianity assumption, such as non-Gaussian continuous, discrete or mixed data. In Section \ref{sec:BDgraph} we describe the main functions implemented in the \pkg{BDgraph} package. In addition, we explain the user interface and the performance of the package by a simple simulation example.
%Section \ref{sec:simulation study} includes an extensive simulation study to evaluate the performance of the Bayesian methods implemented in the \pkg{BDgraph}, as well as to compare them with alternative approaches which are available in the \proglang{R} package \pkg{huge}.
In Section \ref{sec:real data}, using the functions implemented in the \pkg{BDgraph} package, we study two actual datasets. 

% - - - - - - - - - - - - - - - - - - - - - - - - - - - - - - - - - - - - - - - - - - - - - - - - - - - - - - - - - - - - - - - - - - - - - - - - - - - - - - - - - - - - |
\section{User interface}
\label{sec:get started}

In the \proglang{R} environment, one can access and load the \pkg{BDgraph} package by using the following commands:
\begin{Sinput}
R> install.packages( "BDgraph" )
R> library( "BDgraph" )
\end{Sinput}
By loading the \pkg{BDgraph}  package we automatically load the \pkg{igraph} \citep{igraph} package, since the \pkg{BDgraph} package depends on this package for graph visualization. The \pkg{igraph} package is available on the Comprehensive \proglang{R} Archive Network (CRAN) at \url{http://CRAN.R-project.org}.
%We use the \pkg{igraph} package for graph visualization.

To speed up computations, we efficiently implement the \pkg{BDgraph} package by linking the \proglang{C++} code to \proglang{R}. The computationally extensive tasks of the package are implemented in parallel in \proglang{C++} using \pkg{OpenMP} \citep{openmp08}. For the \proglang{C++} code, we use the highly optimized \pkg{LAPACK} \citep{laug} and \pkg{BLAS} \citep{lawson1979basic} linear algebra libraries on systems that provide them. The use of these libraries significantly improves program speed.

We design the \pkg{BDgraph} package to provide a Bayesian framework for undirected graph estimation of different types of datasets such as continuous, discrete or mixed data. The package facilitates a pipeline for analysis by three functional modules; see Figure \ref{fig:modules}. These modules are as follows:
\begin{figure} [!ht]
    \centering
       \includegraphics[width=0.95\textwidth]{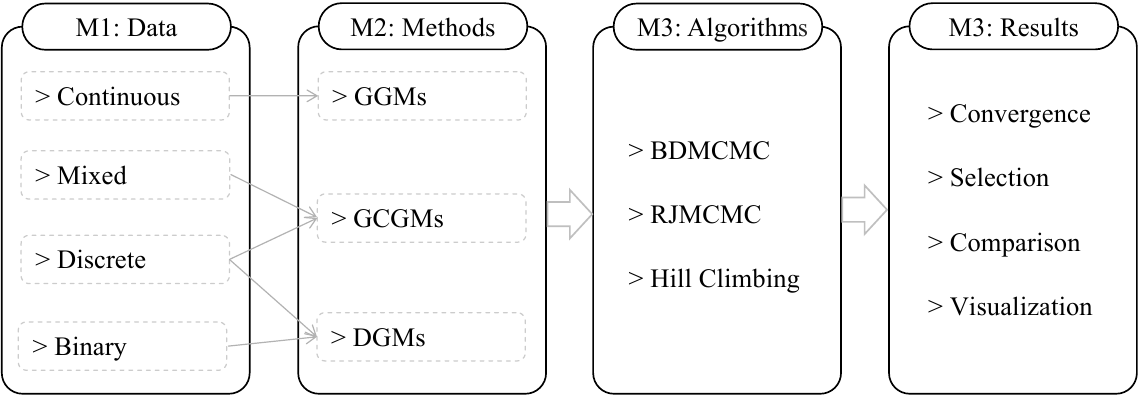}
\caption{
Configuration of the \pkg{BDgraph} package which includes three main parts: (M1) data simulation, (M2) several statistical methods, (M3) several search algorithms, (M4) various functions to evaluate convergence of the search algorithms, estimation of the true graph, assessment and comparison of the results and graph visualization. 
}
\label{fig:modules}
\end{figure}

\textbf{Module 1. Data simulation:} Function \code{bdgraph.sim} simulates multivariate Gaussian, discrete, binary, and mixed data with different undirected graph structures, including \code{"random"}, \code{"cluster"}, \code{"scale-free"}, \code{"lattice"}, \code{"hub"}, \code{"star"}, \code{"circle"}, \code{"AR(1)"}, \code{"AR(2)"}, and \code{"fixed"} graphs. Users can determine the sparsity of the graph structure and can generate mixed data, including \code{"count"}, \code{"ordinal"}, \code{"binary"}, \code{"Gaussian"} and \code{"non-Gaussian"} variables. 

\textbf{Module 2. Methods:} The function \code{bdgraph} and \code{bdgraph.mpl}  provide several estimation methods regarding to the type of data:
\begin{itemize}[noitemsep]
 \item Bayesian graph estimation for the multivariate data that follow the Gaussianity assumption, based on the Gaussian graphical models (GGMs); see \cite{mohammadi2015bayesianStructure, dobra2011bayesian}.
 \item Bayesian graph estimation for multivariate non-Gaussian, discrete, and mixed data, based on Gaussian copula graphical models (GCGMs); see \cite{mohammadi2016bayesian, dobra2011copula}. 
\item Bayesian graph estimation for multivariate discrete and binary data, based on discrete graphical models (DGMs); see \cite{dobra2018}.  
\end{itemize}

\textbf{Module 3. Algorithms:}  The function \code{bdgraph} and \code{bdgraph.mpl} provide several sampling algorithms:
\begin{itemize}[noitemsep]
 \item Birth-death MCMC (BDMCMC) sampling algorithms (Algorithms \ref{algorithm:BDMCMC-GGMs} and \ref{algorithm:BDMCMC-GCGMs}) desciribed in \cite{mohammadi2015bayesianStructure}.
 \item Reversible jump MCMC (RJMCMC) sampling algorithms desciribed in \cite{dobra2011copula}. 
\item Hill-climbing (HC) search algorithm desciribed in \cite{pensar2017marginal}.  
\end{itemize}

\textbf{Module 4. Results:} Includes four types of functions:
\begin{itemize}[noitemsep]
 \item \texttt{Graph selection}: The functions \code{select}, \code{plinks}, and \code{pgraph} provide the selected graph, the posterior link inclusion probabilities and the posterior probability of each graph, respectively. 
 \item \texttt{Convergence check}: The functions \code{plotcoda} and \code{traceplot} provide several visualization plots to monitor the convergence of the sampling algorithms.
 \item \texttt{Comparison and goodness-of-fit}: The functions \code{compare} and \code{plotroc} provide several comparison measures and an ROC plot for  model comparison. 
 \item \texttt{Visualization}: The plotting functions \code{plot.bdgraph} and \code{plot.sim} provide visualizations of the simulated data and estimated graphs. 
 \end{itemize}

% - - - - - - - - - - - - - - - - - - - - - - - - - - - - - - - - - - - - - - - - - - - - - - - - - - - - - - - - - - - - - - - - - - - - - - - - - - - - - - - - - - - - |
\section{Methodological background}
\label{sec:method}

In Section \ref{subsec:GGMs}, we briefly explain the Gaussian graphical model for multivariate data. Then we illustrate the birth-death MCMC algorithm for sampling from the joint posterior distribution over Gaussian graphical models; for more details see \cite{mohammadi2015bayesianStructure}. In Section \ref{subsec:GCGMs}, we briefly describe the Gaussian copula graphical model \citep{dobra2011copula}, which can deal with non-Gaussian, discrete or mixed data. Then we explain the birth-death MCMC algorithm which is designed for the Gaussian copula graphical models; for more details see \cite{mohammadi2016bayesian}.

% - - - - - - - - - - - - - - - - - - - - - - - - - - - - - - - - - - - - - - - - - - - - - - - - - - - - - - - - - - - - - - - - - - - - - - - - - - - - - - - - - - - - |
\subsection{Bayesian Gaussian graphical models}
\label{subsec:GGMs}

In graphical models, each random variable is associated with a node and conditional dependence relationships among random variables are presented as a graph $G=(V,E)$ in which $V = \{1, 2, . . . , p\}$ specifies a set of nodes and a set of existing links $E \subset V\times V$ \citep{lauritzen1996graphical}. Our focus here is on undirected graphs, in which $(i,j) \in E \Leftrightarrow (j,i) \in E$.  The absence of a link between two nodes specifies the pairwise conditional independence of those two variables given the remaining variables, while a link between two variables determines their conditional dependence.

In Gaussian graphical models (GGMs), we assume that the observed data follow multivariate Gaussian distribution $\mathcal{N}_p (\mu,K^{-1})$. Here we assume $\mu=0$. Let $\mathbf{Z} = (Z^{(1)}, ..., Z^{(n)})^\top$ be the observed data of $n$ independent samples, then the likelihood function is 
\begin{eqnarray}
\label{likelihood}
Pr( \mathbf{Z} | K,G) \propto |K|^{n/2} \exp \left\{ -\frac{1}{2} \mbox{tr}(KU) \right\},
\end{eqnarray}
where $ U = \mathbf{Z}^\top \mathbf{Z}$.

In GGMs, conditional independence is implied by the form of the precision matrix. Based on the pairwise Markov property, variables $i$ and $j$ are conditionally independent given the remaining variables, if and only if $K_{ij} = 0$. This property implies that the links in graph $G=(V,E)$ correspond with the nonzero elements of the precision matrix $K$; this means that $E=\{ (i,j) | K_{ij} \neq 0 \}$. Given graph $G$, the precision matrix $K$ is constrained to the cone $\mathbb{P}_{G}$ of symmetric positive definite matrices with elements $K_{ij}$ equal to zero for all $(i,j) \notin E $.

We consider the G-Wishart distribution $W_G(b,D)$ to be a prior distribution for the precision matrix $K$ with density
\begin{eqnarray}
\label{G-Wishart distribution}
Pr(K|G)=\frac{1}{I_G (b,D)} |K|^{(b-2)/2} \exp \left\{ -\frac{1}{2} \mbox{tr}(DK) \right\} 1(K \in \mathbb{P}_{G}),
\end{eqnarray}
where $b > 2$ is the degrees of freedom, $D$ is a symmetric positive definite matrix, $I_G (b,D)$ is the normalizing constant with respect to the graph $G$ and $1(x)$ evaluates to $1$ if $x$ holds, and otherwise to $0$. The G-Wishart distribution is a well-known prior for the precision matrix, since it represents the conjugate prior for multivariate Gaussian data as in \eqref{likelihood}. 

For full graphs, the G-Wishart distribution reduces to the standard Wishart distribution, hence the normalizing constant has an explicit form \citep{muirhead1982aspects}. Also, for decomposable graphs, the normalizing constant has an explicit form \citep{roverato2002hyper}; however, for non-decomposable graphs, it does not. In that case it can be estimated by using the Monte Carlo method \citep{atay2005monte}, the Laplace approximation \citep{lenkoski2011computational}, or recent approximation by \citet{mohammadi2017ratio}. In the \pkg{BDgraph} package, we design the \code{gnorm} function to estimate the log of the normalizing constant by using the Monte Carlo method proposed \cite{atay2005monte}.

Since the G-Wishart prior is a conjugate prior to the likelihood (\ref{likelihood}), the posterior distribution of $K$ is
\begin{eqnarray*}
Pr(K|\mathbf{Z},G) = \frac{1}{I_G (b^*,D^*)} |K|^{(b^*-2)/2} \exp \left\{ -\frac{1}{2} \mbox{tr}(D^*K) \right\},
\end{eqnarray*}
where $b^*=b+n$ and $D^*=D+S$, that is, $W_G(b^*,D^*)$.

% - - - - - - - - - - - - - - - - - - - - - - - - - - - - - - - - - - - - - - - - - - - - - - - - - - - - - - - - - - - - - - - - - - - - - - - - - - - - - - - - - - - - |
\subsubsection{Direct sampler from G-Wishart}
\label{Sample gwishart}

Several sampling methods from the G-Wishart distribution have been proposed; to review existing methods see \citet{wang2012efficient}. More recently, \citet{lenkoski2013direct} has developed an exact sampling algorithm for the G-Wishart distribution, borrowing an idea from \citet{hastie2009elements}.
% - - - - - - - - - - - - - - - - - - - - - - - - - - - - - - - - - - - - - - - - - - - - - - - - - - - - - - - - - - - - - - - - - - - - - - - - - - - - - - - - - - - - |
\begin{algorithm}[!ht]
\renewcommand{\algorithmicrequire}{\textbf{Input:}}
\renewcommand{\algorithmicensure}{\textbf{Output:}}
\caption{. Exact sampling from the precision matrix}
\label{algorithm:sample K}
\begin{algorithmic}[1]
\REQUIRE A graph $G=(V,E)$ with precision matrix $K$ and $\Sigma = K^{-1}$
\ENSURE An exact sample from the precision matrix.
\STATE Set $\Omega = \Sigma$
\REPEAT
  \FOR {$i = 1, ..., p$}
    \STATE Let $N_i \subset V$ be the neighbor set of node $i$ in $G$. Form $\Omega_{N_i}$ and $\Sigma_{N_i,i}$ and solve $\hat{ \beta_{i}^{*} } = \Omega^{-1}_{N_i} \Sigma_{N_i, i}$
    \STATE Form $\hat{ \beta_{i} } \in R^{p-1}$ by padding the elements of $\hat{ \beta_{i}^{*} }$ to the appropriate locations and zeros in those locations not connected to $i$ in $G$
    \STATE Update $\Omega_{i,-i}$ and $\Omega_{-i,i}$ with $\Omega_{-i,-i}\hat{ \beta_{i} }$
  \ENDFOR
\UNTIL{ convergence }  
\RETURN $K = \Omega^{-1}$
\end{algorithmic}
\end{algorithm}
% - - - - - - - - - - - - - - - - - - - - - - - - - - - - - - - - - - - - - - - - - - - - - - - - - - - - - - - - - - - - - - - - - - - - - - - - - - - - - - - - - - - - |

In the \pkg{BDgraph} package, we use Algorithm \ref{algorithm:sample K} to sample from the posterior distribution of the precision matrix. We implement the algorithm in the package as a function \code{rgwish}; see the \proglang{R} code below for illustration.
\vspace{-0.5 \baselineskip} 
\begin{Sinput}
R> adj <- matrix( c( 0, 0, 1, 0, 0, 0, 1, 0, 0 ), 3, 3 )
R> adj    
\end{Sinput}
\vspace{-1 \baselineskip} 
\begin{CodeOutput}
     [,1] [,2] [,3]
[1,]    0    0    1
[2,]    0    0    0
[3,]    1    0    0
\end{CodeOutput}
\vspace{-1 \baselineskip} 
\begin{Sinput}
R> sample <- rgwish( n = 1, adj = adj, b = 3, D = diag( 3 ) )
R> round( sample, 2 )
\end{Sinput}
\vspace{-1 \baselineskip} 
\begin{CodeOutput}
      [,1] [,2]  [,3]
[1,]  2.37 0.00 -2.12
[2,]  0.00 6.15  0.00
[3,] -2.12 0.00  7.26
\end{CodeOutput}
This matrix is a sample from a G-Wishart distribution with $b=3$ and $D=I_3$ as an identity matrix and a graph structure with adjacency matrix \code{adj}.

% - - - - - - - - - - - - - - - - - - - - - - - - - - - - - - - - - - - - - - - - - - - - - - - - - - - - - - - - - - - - - - - - - - - - - - - - - - - - - - - - - - - - |
\subsubsection{BDMCMC algorithm for GGMs}
\label{subsubsec:BDMCMC for GGMs}

Consider the joint posterior distribution of the graph $G$ and the precision matrix $K$ given by
\begin{eqnarray}
\label{joint posterior}
Pr(K, G \mid \mathbf{Z}) \propto Pr(\mathbf{Z} \mid K) ~ Pr( K \mid G) ~ Pr(G).
\end{eqnarray}
For the prior distribution of the graph $G=(V,E)$, we consider a Bernoulli prior on each link inclusion indicator variable as follow
\begin{eqnarray}
\label{graph prior}
Pr(G) \propto \left(  \frac{\theta}{1-\theta}    \right)^{|E|},
\end{eqnarray}
where $|E|$ indicate the number of links in the graph $G$ (graph size) and parameter $\theta \in (0,1)$ is a prior probability of existing link. For the case $\theta = 0.5$ (as a default option of the \pkg{BDgraph}), we will have a uniform distribution over all graph space, as a non-informative prior. For the prior distribution of the precision matrix conditional on the graph $G$, we use a G-Wishart $W_G(b,D)$. 

Here we consider a computationally efficient birth-death MCMC sampling algorithm proposed by \cite{mohammadi2015bayesianStructure} for Gaussian graphical models. The algorithm is based on a continuous time birth-death Markov process, in which the algorithm explores the graph space by adding/removing a link in a birth/death event.

In the birth-death process, for a particular pair of graph $G=(V,E)$ and precision matrix $K$, each link dies independently of the rest as a Poisson process with death rate $\delta_e(K)$. Since the links are independent, the overall death rate is $\delta(K) = \sum_{e \in E}\delta_e(K)$. Birth rates $\beta_e(K)$ for $e \notin E$ are defined similarly. Thus the overall birth rate is $\beta(K) = \sum_{e \notin E}\beta_e(K)$.

Since the birth and death events are independent Poisson processes, the time between two successive events is exponentially distributed with mean $1/(\beta(K)+\delta(K))$. The time between successive events can be considered as inverse support for any particular instance of the state $(G,K)$. The probabilities of birth and death events are
\begin{eqnarray}
\label{prob.birth}
Pr(\mbox{birth of link  } e)= \frac{\beta_{e}(K)}{\beta(K)+\delta(K)}, \qquad \mbox{for each} \ \ e \notin E,
\end{eqnarray}
\begin{eqnarray}
\label{prob.death}
Pr(\mbox{death of link  } e)= \frac{\delta_{e}(K)}{\beta(K)+\delta(K)}, \qquad \mbox{for each} \ \ e \in E.
\end{eqnarray}

The birth and death rates of links occur in continuous time with the rates determined by the stationary distribution of the process. The BDMCMC algorithm is designed in such a way that the stationary distribution is equal to the target joint posterior distribution of the graph and the precision matrix \eqref{joint posterior}. 

\citet[Theorem 3.1]{mohammadi2015bayesianStructure} derived a condition that guarantees the above birth and death process converges to our target joint posterior distribution \eqref{joint posterior}. By following their Theorem we define the birth and death rates, as below
% \citet[Theorem 3.1]{mohammadi2015bayesianStructure} prove 
% the BDMCMC sampling algorithm converges to the target joint posterior distribution of the graph and the precision matrix,
% by considering the birth and death rates as ratios of joint posterior distributions, as below
\begin{equation}
\label{birthrate}
\beta_{e}(K) = \min \left\{ \frac{Pr(G^{+e},K^{+e} |\mathbf{Z})}{Pr(G,K |\mathbf{Z})}, 1 \right\}, \ \ \mbox{for each} \ \ e \notin E,
\end{equation}
\begin{equation}
\label{deathrate}
\delta_{e}(K) = \min \left\{ \frac{Pr(G^{-e},K^{-e} |\mathbf{Z})}{Pr(G,K |\mathbf{Z} )}, 1 \right\}, \ \ \mbox{for each} \ \ e \in E,
\end{equation}
in which $G^{+e}=(V, E \cup \{ e \})$ and $K^{+e} \in \mathbb{P}_{G^{+e}}$ and similarly $G^{-e} = ( V, E \setminus \{ e \})$ and $K^{-e} \in \mathbb{P}_{G^{-e}}$. For computation part related to the ratio of posterior see \cite{mohammadi2017ratio}.  

Algorithm \ref{algorithm:BDMCMC-GGMs} provides the pseudo-code for our BDMCMC sampling scheme which is based on the above birth and death rates. 
% Based on the above rates we determine our BDMCMC sampling algorithm as follows: 
% - - - - - - - - - - - - - - - - - - - - - - - - - - - - - - - - - - - - - - - - - - - - - - - - - - - - - - - - - - - - - - - - - - - - - - - - - - - - - - - - - - - - |
\begin{algorithm}
\renewcommand{\algorithmicrequire}{\textbf{Input:}}
\renewcommand{\algorithmicensure}{\textbf{Output:}} 
\caption{. BDMCMC algorithm for GGMs}
\label{algorithm:BDMCMC-GGMs}
\begin{algorithmic}[1]
\REQUIRE A graph $G=(V,E)$ and a precision matrix $K$.
\ENSURE Samples from the joint posterior distribution of $(G,K)$, \eqref{joint posterior}, and waiting times.
\FOR {N iteration}
    \STATE \textbf{1. Sample from the graph.} Based on birth and death process
    \STATE $\; \; \; $ 1.1. Calculate the birth rates by (\ref{birthrate}) and $\beta(K)= \sum_{e \in \notin E}{\beta_{e}(K)}$
    \STATE $\; \; \; $ 1.2. Calculate the death rates by (\ref{deathrate}) and $\delta(K)=\sum_{e \in E}{\delta_{e}(K)}$
    \STATE $\; \; \; $ 1.3. Calculate the waiting time by $W(K)= 1/(\beta(K)+\delta(K))$
    \STATE $\; \; \; $ 1.4. Simulate the type of jump (birth or death) by (\ref{prob.birth}) and (\ref{prob.death})
  \STATE  \textbf{2. Sample from the precision matrix.} By using Algorithm \ref{algorithm:sample K}.
\ENDFOR
\end{algorithmic}
\end{algorithm}
% - - - - - - - - - - - - - - - - - - - - - - - - - - - - - - - - - - - - - - - - - - - - - - - - - - - - - - - - - - - - - - - - - - - - - - - - - - - - - - - - - - - - |

Note, step $1$ of the algorithm is suitable for parallel computation. In the \pkg{BDgraph}, we implement this step of algorithm in parallel using \pkg{OpenMP} in \proglang{C++} to speed up the computations.

\begin{figure} [!ht] % [!ht]
\centering
\includegraphics[width=1\textwidth]{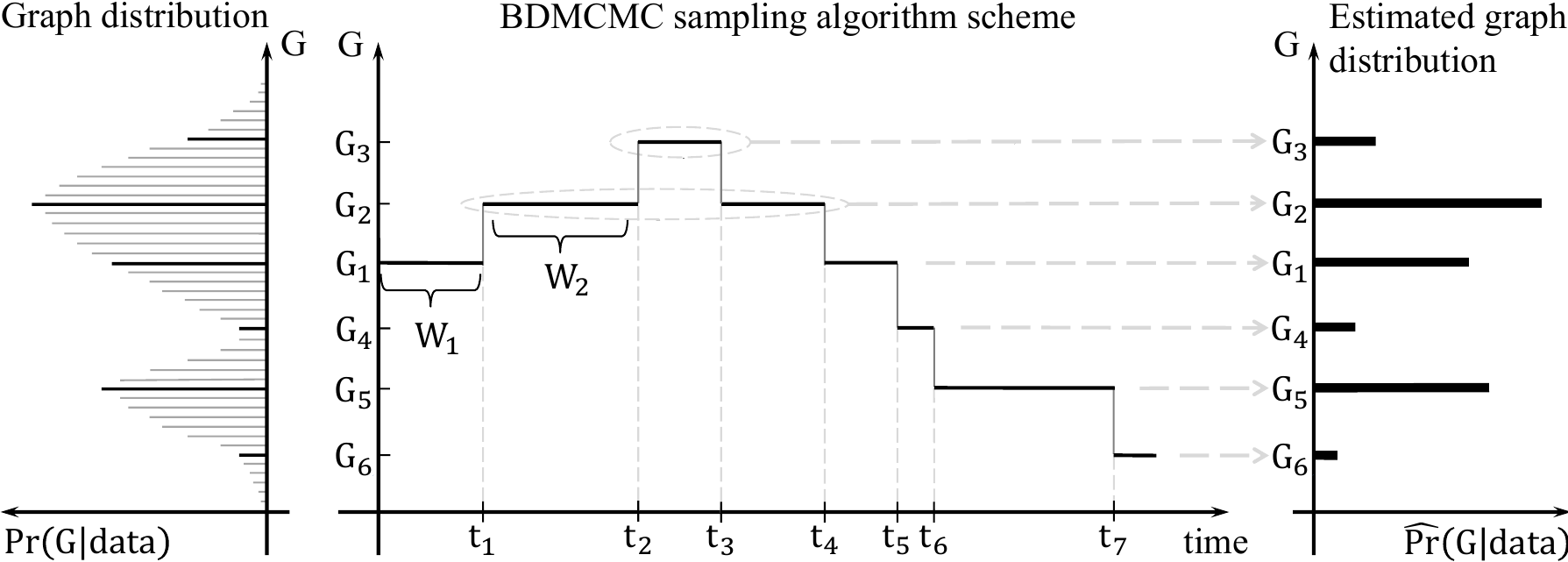}
\caption{ \label{fig:BDMCMC} This image visualizes the Algorithm \ref{algorithm:BDMCMC-GGMs}.   The left side shows the true posterior distribution of the graph. The middle panel presents a continuous time BDMCMC sampling algorithm where $ \left\{ W_1, W_2,... \right\}$ denote waiting times and $ \left\{ t_1, t_2,... \right\}$ denote jumping times. The right side denotes the estimated posterior probability of the graphs in proportion to the total of their waiting times, according to the Rao-Blackwellized estimator. }
\end{figure}

The BDMCMC sampling algorithm is designed in such a way that a sample $(G,K)$ is obtained at certain jump moments, $\{ t_1, t_2, ... \}$ (see Figure \ref{fig:BDMCMC}). For efficient posterior inference of the parameters, we use the Rao-Blackwellized estimator, which is an efficient estimator for continuous time MCMC algorithms \citep[Section 2.5]{cappe2003reversible}. By using the Rao-Blackwellized estimator, for example, one can estimate the posterior distribution of the graphs proportional to the total waiting 
times of each graph.  

% - - - - - - - - - - - - - - - - - - - - - - - - - - - - - - - - - - - - - - - - - - - - - - - - - - - - - - - - - - - - - - - - - - - - - - - - - - - - - - - - - - - - |
\subsection{Gaussian copula graphical models}
\label{subsec:GCGMs}

In practice we encounter both discrete and continuous variables; Gaussian copula graphical modelling has been proposed by \cite{dobra2011copula} to describe dependencies between such heterogeneous variables. Let $\Yb$ (as observed data) be a collection of continuous, binary, ordinal or count variables with the marginal distribution $F_j$ of $Y_j$ and $F_j^{-1}$ as its pseudo inverse. For constructing a joint distribution of $\Yb$, we introduce a multivariate Gaussian latent variable as follows:
\begin{eqnarray}
\label{model: data and latend}
Z_1, ..., Z_n \stackrel{iid}{\sim} \mathcal{N}_p(0, \Gamma(K)), \nonumber \\
Y_{ij} = F_j^{-1}(\Phi(Z_{ij})),
\end{eqnarray}
where $\Gamma(K)$ is the correlation matrix for a given precision matrix $K$. The joint distribution of $\Yb$ is given by
\begin{eqnarray}
\label{jointdf}
  Pr\left(Y_1 \leq Y_1, \ldots, Y_p  \leq Y_p \right) = C( F_1(Y_1), \ldots, F_p(Y_p) \mid \Gamma(K)),  
\end{eqnarray}
where $C(\cdot)$ is the Gaussian copula given by 
\[
 C(u_1, \ldots, u_p \mid \Gamma) = \Phi_p\left(\Phi^{-1}(u_1), \ldots, \Phi^{-1}(u_p) \mid \Gamma\right),
\]
with $u_v = F_v(Y_v)$ and $\Phi_p(\cdot)$ is the cumulative distribution of multivariate Gaussian and $\Phi(\cdot)$ is the cumulative distribution of univariate Gaussian distributions. It follows that $Y_v = F_v^{-1}\left(\Phi(Z_v)\right)$ for $v={1, ... ,p}$. If all variables are continuous then the margins are unique; thus zeros in $K$ imply conditional independence, as in Gaussian graphical models \citep{hoff2007extending, abegaz2015copula}. For discrete variables, the margins are not unique but still well-defined \citep{nelsen2007introduction}.

In semiparametric copula estimation, the marginals are treated as nuisance parameters and estimated by the rescaled empirical distribution. The joint distribution in (\ref{jointdf}) is then parametrized only by the correlation matrix of the Gaussian copula.  We are interested to infer the underlying graph structure of the observed variables $\Yb$ implied by the continuous latent variables $\Zb$.  Since $\Zb$ are unobservable we follow the idea of \citet{hoff2007extending} of associating them with the observed data as below. 

Given the observed data $\Yb$ from a sample of $n$ observations, we constrain the samples from latent variables $\Zb$ to belong to the set 
\begin{eqnarray*}
 \mathcal{D}(\Yb) = \{\Zb \in \mathbb{R}^{n \times p} : L_j^r(\Zb) < z_j^{(r)} < U_j^r(\Zb), r=1, \ldots, n; j=1, \ldots, p  \},
\end{eqnarray*}
where 
\begin{eqnarray}
\label{truncated set}
L_j^r(\Zb) = \max \left\{ Z_j^{(s)}: Y_j^{(s)} < Y_j^{(r)} \right\} \mbox{ and } U_j^r(\Zb) = \min\left\{ Z_j^{(s)}: Y_j^{(r)} < Y_j^{(s)}\right\}.
\end{eqnarray}

Following \cite{hoff2007extending} we infer the latent space by substituting the observed data $\Yb$ with the event $\mathcal{D}(\Yb)$ and define the likelihood as
\begin{eqnarray*}
 Pr(\Yb \mid K, G, F_1,...,F_p) = Pr(\Zb \in \mathcal{D}(\Yb) \mid K, G) ~ Pr(\Yb \mid \Zb \in \mathcal{D}(\Yb), K, G, F_1,...,F_p ).
\end{eqnarray*}
The only part of the observed data likelihood relevant for inference on $K$ is $ Pr(\Zb \in \mathcal{D}(\Yb) \mid K, G)$.  Thus, the likelihood function is given by
\begin{eqnarray}
 Pr(\Zb \in \mathcal{D}(\Yb) \mid K, G) = Pr(\Zb \in \mathcal{D}(\Yb) \mid K, G) = \int_{\mathcal{D}(\Yb)} Pr(\Zb \mid K, G) dZ
\end{eqnarray}
where $Pr(\Zb \mid K, G)$ is defined in (\ref{likelihood}).

% - - - - - - - - - - - - - - - - - - - - - - - - - - - - - - - - - - - - - - - - - - - - - - - - - - - - - - - - - - - - - - - - - - - - - - - - - - - - - - - - - - - - |
\subsubsection{BDMCMC algorithm for GCGMs}
\label{subsubsec:BDMCMC for GCGMs}

The joint posterior distribution of the graph $G$ and precision matrix $K$ for the GCGMs is 
\begin{eqnarray}
\label{joint posterior GCGMs}
Pr(K,G | Z \in \mathcal{D}(\Yb)) \propto Pr(K,G) Pr(Z \in \mathcal{D}(\Yb) | K, G).
\end{eqnarray}
Sampling from this posterior distribution can be done by using the birth-death MCMC algorithm. \cite{mohammadi2016bayesian} have developed and extended the birth-death MCMC algorithm to more general cases of GCGMs. We summarize their algorithm as follows:
% - - - - - - - - - - - - - - - - - - - - - - - - - - - - - - - - - - - - - - - - - - - - - - - - - - - - - - - - - - - - - - - - - - - - - - - - - - - - - - - - - - - - |
\begin{algorithm}
\renewcommand{\algorithmicrequire}{\textbf{Input:}}
\renewcommand{\algorithmicensure}{\textbf{Output:}} 
\caption{. BDMCMC algorithm for GCGMs}
\label{algorithm:BDMCMC-GCGMs}
\begin{algorithmic}[1]
\REQUIRE A graph $G=(V,E)$ and a precision matrix $K$.
\ENSURE Samples from the joint posterior distribution of $(G,K)$, \eqref{joint posterior GCGMs}, and waiting times.
\FOR {N iteration}
\STATE \textbf{1. Sample the latent data.} For each $r \in V$ and $j \in \{ 1,...,n \}$, we update the latent values from its full conditional distribution as follows
 $$Z_{r}^{(j)} | Z_{V \setminus \{ r \} } = z_{V \setminus \{ r \} }^{(j)}, K \sim N (- \sum_{r'} { K_{rr'} z_{r'}^{(j)} / K_{rr} }, 1/K_{rr} ),$$
 truncated to the interval $\left[ L_r^j(\Zb), U_r^j(\Zb) \right]$ in (\ref{truncated set}).
\STATE \textbf{2. Sample from the graph.} Same as step $1$ in Algorithm \ref{algorithm:BDMCMC-GGMs}.

  \STATE \textbf{3. Sample from the precision matrix.} By using Algorithm \ref{algorithm:sample K}.
\ENDFOR
\end{algorithmic}
\end{algorithm}
% - - - - - - - - - - - - - - - - - - - - - - - - - - - - - - - - - - - - - - - - - - - - - - - - - - - - - - - - - - - - - - - - - - - - - - - - - - - - - - - - - - - - |

In step $1$, the latent variables $\Zb$ are sampled conditional on the observed data $\Yb$. The other steps are the same as in Algorithm \ref{algorithm:BDMCMC-GGMs}.

\textbf{Remark:} in cases where all variables are continuous, we do not need to sample from latent variables in each iteration of Algorithm \ref{algorithm:BDMCMC-GGMs}, since all margins in the Gaussian copula are unique. Thus, for these cases, we transfer our non-Gaussian data to Gaussian, and then we run Algorithm \ref{algorithm:BDMCMC-GGMs}; see example \ref{subsec:gene data}.

% - - - - - - - - - - - - - - - - - - - - - - - - - - - - - - - - - - - - - - - - - - - - - - - - - - - - - - - - - - - - - - - - - - - - - - - - - - - - - - - - - - - - |
\subsubsection{ Alternative RJMCMC algorithm }
\label{subsubsec:RJMCMC}

RJMCMC is a special case of the trans-dimensional MCMC methodology \citet{green2003trans}. The RJMCMC approach is based on an ergodic discrete-time Markov chain. In graphical models, a RJMCMC algorithm can be designed in such a way that its stationary distribution is the joint posterior distribution of the graph and the parameters of the graph, e.g., \ref{joint posterior} for GGMs and \ref{joint posterior GCGMs} for GCGMs.

A RJMCMC can be implemented in various different ways. \cite{giudici1999decomposable} implemented this algorithm only for the decomposable GGMs, because of the expensive computation of the normalizing constant $I_G(b,D)$. The RJMCMC approach developed by \cite{dobra2011bayesian} and \cite{dobra2011copula} is based on the Cholesky decomposition of the precision matrix. It uses an approximation for dealing with the extensive computation of the normalizing constant. To avoid the intractable normalizing constant calculation, \citet{lenkoski2013direct} and \citet{wang2012efficient} implemented a special case of RJMCMC algorithm, which is based on the exchange algorithm \citep{murray2012mcmc}. Our implementation of RJMCMC algorithm in the \pkg{BDgraph} package defines the acceptance probability proportional to the birth/death rates in our BDMCMC algorithm. Moreover, we implement the exact sampling of G-Wishart distribution, as described in Section \ref{Sample gwishart}. Besides, we using the result of \cite{mohammadi2017ratio} for the ratio of the normalizing constant of G-Wishart distribution.

% - - - - - - - - - - - - - - - - - - - - - - - - - - - - - - - - - - - - - - - - - - - - - - - - - - - - - - - - - - - - - - - - - - - - - - - - - - - - - - - - - - - - |
\section{ The BDgraph environment }
\label{sec:BDgraph}

The \pkg{BDgraph} package provides a set of comprehensive tools related to Bayesian graphical models; we describe below the essential functions available in the package. 

% - - - - - - - - - - - - - - - - - - - - - - - - - - - - - - - - - - - - - - - - - - - - - - - - - - - - - - - - - - - - - - - - - - - - - - - - - - - - - - - - - - - - |
\subsection{ Posterior sampling }
\label{subsec:bdgraph}

We design the function \code{bdgraph}, as the main function of the package, to take samples from the posterior distributions based on both of our Bayesian frameworks (GGMs and GCGMs). By default, the \code{bdgraph} function is based on underlying sampling algorithms (Algorithms \ref{algorithm:BDMCMC-GGMs} and \ref{algorithm:BDMCMC-GCGMs}). Moreover, as an alternative to those BDMCMC sampling algorithms, we implement RJMCMC sampling algorithms for both the Gaussian and non-Gaussian frameworks. By using the following function
\begin{Sinput}
bdgraph( data, n = NULL, method = "ggm", algorithm = "bdmcmc", iter = 5000, 
  burnin = iter / 2, not.cont = NULL, g.prior = 0.5, df.prior = 3,
  g.start = "empty", jump = NULL, save = FALSE, print = 1000, cores = NULL,
  threshold = 1e-8 )
\end{Sinput}
we obtain a sample from our target joint posterior distribution. \code{bdgraph} returns an object of \code{S3} class type ``\code{bdgraph}''. The functions \code{plot}, \code{print} and \code{summary} are working with the object  ``\code{bdgraph}''. The input \code{data} can be an ($n \times p$) \code{matrix} or a \code{data.frame} or a covariance ($p \times p$) matrix ($n$ is the sample size and $p$ is the dimension); it can also be an object of class ``\code{sim}'', which is the output of function \code{bdgraph.sim}.

The argument \code{method} determines the type of methods, GGMs, GCGMs. Option ``\code{ggm}'' is based on Gaussian graphical models (Algorithm \ref{algorithm:BDMCMC-GGMs}) that is designed for multivariate Gaussian data. Option ``\code{gcgm}'' is based on the GCGMs (Algorithm \ref{algorithm:BDMCMC-GCGMs}) that is designed for non-Gaussian data such as, non-Gaussian continuous, discrete or mixed data.

The argument \code{algorithm} refers the type of sampling algorithms which could be based on BDMCMC or RJMCMC. Option ``\code{bdmcmc}'' (as default) is for the BDMCMC sampling algorithms (Algorithms \ref{algorithm:BDMCMC-GGMs} and \ref{algorithm:BDMCMC-GCGMs}). Option ``\code{rjmcmc}'' is for the RJMCMC sampling algorithms, which are alternative algorithms. See \citet[Section 4]{mohammadi2015bayesianStructure}, \citet[Section 2.2.3]{mohammadi2016bayesian}. %  and the Supplementary material for evaluation of these algorithms. 
%See Section \ref{sec:simulation study} to evaluate the performance of these algorithms.

The argument \code{g.start} specifies the initial graph for our sampling algorithm. It could be \code{empty} (default) or \code{full}. Option \code{empty} means the initial graph is an empty graph and \code{full} means a full graph. It also could be an object with \code{S3} class \code{"bdgraph"}, which allows users to run the sampling algorithm from the last objects of the previous run.     

The argument \code{jump} determines the number of links that are simultaneously updated in the BDMCMC algorithm. 

For parallel computation in \proglang{C++} which is based on  \pkg{OpenMP} \citep{openmp08}, user can use argument \code{cores} which specifies the number of cores to use for parallel execution. 

Note, the package \pkg{BDgraph} has two other sampling functions, \code{bdgraph.mpl} and \code{bdgraph.ts} which are designed in the similar framework as the function \code{bdgraph}. The function \code{bdgraph.mpl} is for Bayesian model determination in undirected graphical models based on marginal pseudo-likelihood, for both continuous and discrete variables; For more details see \cite{dobra2018}. The function \code{bdgraph.ts} is for Bayesian model determination in time series graphical models \citep{tank2015bayesian}. 

% - - - - - - - - - - - - - - - - - - - - - - - - - - - - - - - - - - - - - - - - - - - - - - - - - - - - - - - - - - - - - - - - - - - - - - - - - - - - - - - - - - - - |
\subsection{ Posterior graph selection }
\label{subsec:selection}

We design the \pkg{BDgraph} package in such a way that posterior graph selection can be done based on both Bayesian model averaging (BMA), as default, and maximum a posterior probability (MAP). The functions \code{select} and \code{plinks} are designed for the objects of class \code{bdgraph} to provide BMA and MAP estimations for posterior graph selection. 

The function
\begin{Sinput}
plinks( bdgraph.obj, round = 2, burnin = NULL )
\end{Sinput}
provides estimated posterior link inclusion probabilities for all possible links, which is based on BMA estimation. In cases where the sampling algorithm is based on BDMCMC, these probabilities for all possible links $e=(i,j)$ in the graph can be estimated using a Rao-Blackwellized estimate \citep[Section 2.5]{cappe2003reversible} based on
\begin{eqnarray}
\label{posterior-link}
 Pr( e \in E | data )= \frac{\sum_{t=1}^{N}{1(e \in E^{(t)}) W(K^{(t)}) }}{\sum_{t=1}^{N}{W(K^{(t)})}},
\end{eqnarray}
where $N$ is the number of iteration and $W(K^{(t)})$ are the weights of the graph $G^{(t)}$ with the precision matrix $K^{(t)}$.

The function 
\begin{Sinput}
select( bdgraph.obj, cut = NULL, vis = FALSE )
\end{Sinput}
provides the inferred graph based on both BMA (as default) and MAP estimators. The inferred graph based on BMA estimation is a graph with links for which the estimated posterior probabilities are greater than a certain cut-point (as default \code{cut=0.5}). The inferred graph based on MAP estimation is a graph with the highest posterior probability.

Note, for posterior graph selection based on MAP estimation we should save all adjacency matrices by using the option \code{save = TRUE} in the function \code{bdgraph}. Saving all the adjacency matrices could, however, cause memory problems; to see how we cope with this problem the reader is referred to Appendix \ref{appendix}.

% - - - - - - - - - - - - - - - - - - - - - - - - - - - - - - - - - - - - - - - - - - - - - - - - - - - - - - - - - - - - - - - - - - - - - - - - - - - - - - - - - - - - |
\subsection{ Convergence check }
\label{subsec:plotcoda}

In general, convergence in MCMC approaches can be difficult to evaluate. From a theoretical point of view, the sampling distribution will converge to the target joint posterior distribution as the number of iteration increases to infinity. Because we normally have little theoretical insight about how quickly MCMC algorithms converge to the target stationary distribution we therefore rely on post hoc testing of the sampled output. In general, the sample is divided into two parts: a ``burn-in'' part of the sample and the remainder, in which the chain is considered to have converged sufficiently close to the target posterior distribution. Two questions then arise: How many samples are sufficient? How long should the burn-in period be? 

The \code{plotcoda} and \code{traceplot} are two visualization functions for the objects of class \code{bdgraph} that make it possible to check the convergence of the search algorithms in \pkg{BDgraph}. The function
\begin{Sinput}
plotcoda( bdgraph.obj, thin = NULL, control = TRUE, main = NULL, ... )
\end{Sinput}
provides the trace of estimated posterior probability of all possible links to check convergence of the search algorithms. Option \code{control} is designed for the case where if \code{control=TRUE} (as default) and the dimension ($p$) is greater than $15$, 
then $100$ links are randomly selected for visualization.

The function
\begin{Sinput}
traceplot( bdgraph.obj, acf = FALSE, pacf = FALSE, main = NULL, ... )
\end{Sinput}
provides the trace of graph size to check convergence of the search algorithms. Option \code{acf} is for visualization of the autocorrelation functions for graph size; option \code{pacf} visualizes the partial autocorrelations.

% - - - - - - - - - - - - - - - - - - - - - - - - - - - - - - - - - - - - - - - - - - - - - - - - - - - - - - - - - - - - - - - - - - - - - - - - - - - - - - - - - - - - |
\subsection{ Comparison and goodness-of-fit }
\label{subsec:compare}

The functions \code{compare} and \code{plotroc} are designed to evaluate and compare the performance of the selected graph. These functions are particularly useful for simulation studies. With the function 
\begin{Sinput}
compare( target, est, est2 = NULL, est3 = NULL, est4 = NULL, main = NULL,
  vis = FALSE )
\end{Sinput}
we can evaluate the performance of the Bayesian methods available in our \pkg{BDgraph} package and compare them with alternative approaches. This function provides several measures such as the balanced $F$-score measure \citep{baldi2000assessing}, which is defined as follows:
\begin{eqnarray}
\label{f1}
F_1\mbox{-score} = \frac{2 \mbox{TP}}{2 \mbox{TP + FP + FN}},
\end{eqnarray}
where TP, FP and FN are the number of true positives, false positives and false negatives, respectively. The $F_1$-score lies between $0$ and $1$, where $1$ stands for perfect identification and $0$ for no true positives.

The function 
\begin{Sinput}
plotroc( target, est, est2 = NULL, est3 = NULL, est4 = NULL, cut = 20, 
  smooth = FALSE, label = TRUE, main = "ROC Curve" )
\end{Sinput}
provides a ROC plot for visualization comparison based on the estimated posterior link inclusion probabilities. 

% - - - - - - - - - - - - - - - - - - - - - - - - - - - - - - - - - - - - - - - - - - - - - - - - - - - - - - - - - - - - - - - - - - - - - - - - - - - - - - - - - - - - |
\subsection{ Data simulation }
\label{subsec:bdgraph.sim}

The function \code{bdgraph.sim} is designed to simulate different types of datasets with various graph structures. The function
\begin{Sinput}
bdgraph.sim( p = 10, graph = "random", n = 0, type = "Gaussian", prob = 0.2, 
  size = NULL, mean = 0, class = NULL, cut = 4, b = 3, D = diag( p ), 
  K = NULL, sigma = NULL, vis = FALSE )
\end{Sinput}
can simulate multivariate Gaussian, non-Gaussian, discrete, binary and mixed data with different undirected graph structures, including \code{"random"}, \code{"cluster"}, \code{"scale-free"}, \code{"lattice"}, \code{"hub"}, \code{"star"}, \code{"circle"}, \code{"AR(1)"}, \code{"AR(2)"}, and \code{"fixed"} graphs. Users can specify the type of multivariate data by option \code{type} and the graph structure by option \code{graph}. They can determine the sparsity level of the obtained graph by using option \code{prob}. With this function users can generate mixed data from \code{"count"}, \code{"ordinal"}, \code{"binary"}, \code{"Gaussian"} and \code{"non-Gaussian"} distributions. \code{bdgraph.sim} returns an object of the \code{S3} class type ``\code{sim}''. Functions \code{plot} and \code{print} work with this object type.

There is another function in the  \pkg{BDgraph} package with the name \code{graph.sim} which is designed to simulate different types of graph structures. The function
\begin{Sinput}
graph.sim( p = 10, graph = "random", prob = 0.2, size = NULL, class = NULL, 
  cut = 4, vis = FALSE )
\end{Sinput}
can simulate different undirected graph structures, including  \code{"random"}, \code{"cluster"}, \code{"scale-free"}', \code{"lattice"}, \code{"hub"}, \code{"star"}, and \code{"circle"} graphs. Users can specify the type of graph structure by option \code{graph}. They can determine the sparsity level of the obtained graph by using option \code{prob}. \code{bdgraph.sim} returns an object of the \code{S3} class type ``\code{graph}''. Functions \code{plot} and \code{print} work with this object type.

% - - - - - - - - - - - - - - - - - - - - - - - - - - - - - - - - - - - - - - - - - - - - - - - - - - - - - - - - - - - - - - - - - - - - - - - - - - - - - - - - - - - - |
\section{An example on simulated data }
\label{sec:simple example}

We illustrate the user interface of the \pkg{BDgraph} package by use of a simple simulation. We perform all the computations on an MacBook Pro with $2.9$ GHz Intel Core i$7$ processor. By using the function \code{bdgraph.sim} we simulate $60$ observations ($n=60$) from a multivariate Gaussian distribution with $8$ variables ($p=8$) and ``scale-free'' graph structure, as below.
\begin{Sinput}
R> data.sim <- bdgraph.sim( n = 60, p = 8, graph = "scale-free", 
+                           type = "Gaussian" )
R> round( head( data.sim $ data, 4 ), 2 ) 
\end{Sinput}
\begin{CodeOutput}
      [,1]  [,2]  [,3]  [,4]  [,5]  [,6]  [,7]  [,8]
[1,]  0.72 -0.91 -1.23 -0.16  0.20 -0.47  0.08  1.07
[2,]  0.25 -0.11  0.09  0.53  0.10 -0.04 -0.13 -0.67
[3,] -0.42 -0.09 -0.28 -0.42  2.04  0.84 -0.79  1.24
[4,] -0.33 -0.50  0.68 -1.33 -1.15  0.25 -0.35  2.97
\end{CodeOutput}

Since the generated data are Gaussian, we run the BDMCMC algorithm which is based on Gaussian graphical models. For this we choose \code{method = "ggm"}, as follows:
\begin{Sinput}
R> sample.bdmcmc <- bdgraph( bdgraph( data = data.sim, method = "ggm", 
+                            algorithm = "bdmcmc", iter = 5000, save = TRUE )
\end{Sinput}
We choose option ``\code{save = TRUE}'' to save the samples in order to check convergence of the algorithm. Running this function takes less than one second, as the computational intensive tasks are performed in \proglang{C++} and interfaced with \proglang{R}.

Since the function \code{bdgraph} returns an object of class \code{S3}, users can see the summary result as follows
\begin{Sinput}
R> summary( sample.bdmcmc )
\end{Sinput}
\begin{CodeOutput}
$selected_g
     [,1] [,2] [,3] [,4] [,5] [,6] [,7] [,8]
[1,]    0    1    1    0    0    0    1    0
[2,]    0    0    0    1    0    0    0    0
[3,]    0    0    0    0    0    1    0    0
[4,]    0    0    0    0    0    0    0    1
[5,]    0    0    0    0    0    0    0    0
[6,]    0    0    0    0    0    0    0    0
[7,]    0    0    0    0    0    0    0    0
[8,]    0    0    0    0    0    0    0    0

$p_links
     [,1] [,2] [,3] [,4] [,5] [,6] [,7] [,8]
[1,]    0 0.51 1.00 0.27 0.21 0.31 0.74 0.11
[2,]    0 0.00 0.29 1.00 0.25 0.18 0.49 0.14
[3,]    0 0.00 0.00 0.24 0.27 0.79 0.44 0.22
[4,]    0 0.00 0.00 0.00 0.32 0.30 0.34 1.00
[5,]    0 0.00 0.00 0.00 0.00 0.25 0.40 0.22
[6,]    0 0.00 0.00 0.00 0.00 0.00 0.23 0.37
[7,]    0 0.00 0.00 0.00 0.00 0.00 0.00 0.19
[8,]    0 0.00 0.00 0.00 0.00 0.00 0.00 0.00

$K_hat
      [,1]  [,2]  [,3]  [,4]  [,5]  [,6]  [,7]  [,8]
[1,]  3.81  0.33  3.19 -0.09  0.04  0.14 -0.84  0.02
[2,]  0.33  4.24 -0.06 -3.43 -0.07 -0.02  0.41 -0.02
[3,]  3.19 -0.06  5.54 -0.08 -0.06 -0.75  0.41  0.08
[4,] -0.09 -3.43 -0.08  9.28 -0.15  0.10 -0.18  1.62
[5,]  0.04 -0.07 -0.06 -0.15  0.76 -0.06  0.16 -0.04
[6,]  0.14 -0.02 -0.75  0.10 -0.06  3.08  0.04 -0.14
[7,] -0.84  0.41  0.41 -0.18  0.16  0.04  5.56  0.04
[8,]  0.02 -0.02  0.08  1.62 -0.04 -0.14  0.04  1.21

\end{CodeOutput}
The summary results are the adjacency matrix of the selected graph (\code{selected_g}) based on BMA estimation, the estimated posterior probabilities of all possible links (\code{p_links}) and the estimated precision matrix (\code{K_hat}).

\begin{figure} [!ht] % [!ht]
    \centering
    \includegraphics[width=0.7\textwidth]{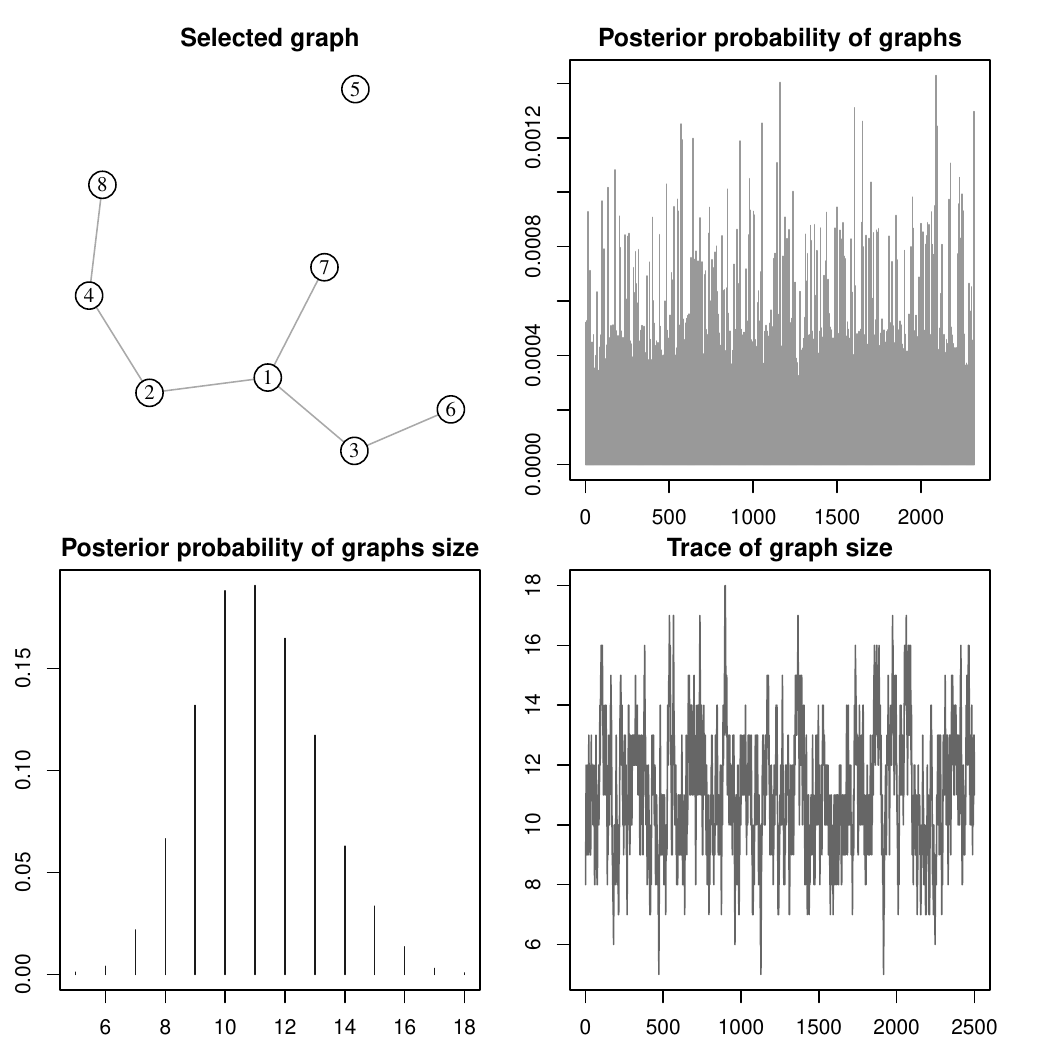}
\caption{
Visualization summary of simulation data based on output of \code{bdgraph} function. The figure at the top-left is the inferred graph with the highest posterior probability. The figure at the top-right gives the estimated posterior probabilities of all visited graphs. The figure at the bottom-left gives the estimated posterior probabilities of all visited graphs based on the size of the graphs. The figure at in the bottom-right is the trace of our algorithm based on the size of the graphs.
}
\label{fig:summary-bdgraph}
\end{figure}

In addition, the function \code{summary} reports a visualization summary of the results as we can see in Figure \ref{fig:summary-bdgraph}. At the top-left is the graph with the highest posterior probability. The plot at the top-right gives the estimated posterior probabilities of all the graphs which are visited by the BDMCMC algorithm; it indicates that our algorithm visits more than $2000$ different graphs. The plot at the bottom-left gives the estimated posterior probabilities of the size of the graphs; it indicates that our algorithm visited mainly graphs with sizes between $4$ and $18$ links. At the bottom-right is the trace of our algorithm based on the size of the graphs.

The function \code{compare} provides several measures to evaluate the performance of our algorithms and compare them with alternative approaches with respect to the true graph structure. To evaluate the performance of our BDMCMC algorithm (Algorithm \ref{algorithm:BDMCMC-GGMs}) and compare it with that of an alternative algorithm, we also run the RJMCMC algorithm under the same conditions as below. 
\begin{Sinput}
R> sample.rjmcmc <- bdgraph( data = data.sim, method = "ggm", 
+                            algorithm = "rjmcmc", iter = 5000, save = TRUE )
\end{Sinput}
where the sampling algorithm from the joint posterior distribution is based on the RJMCMC algorithm. 

\begin{figure} [!ht] % [!ht]
    \centering
    \includegraphics[width=0.45\textwidth]{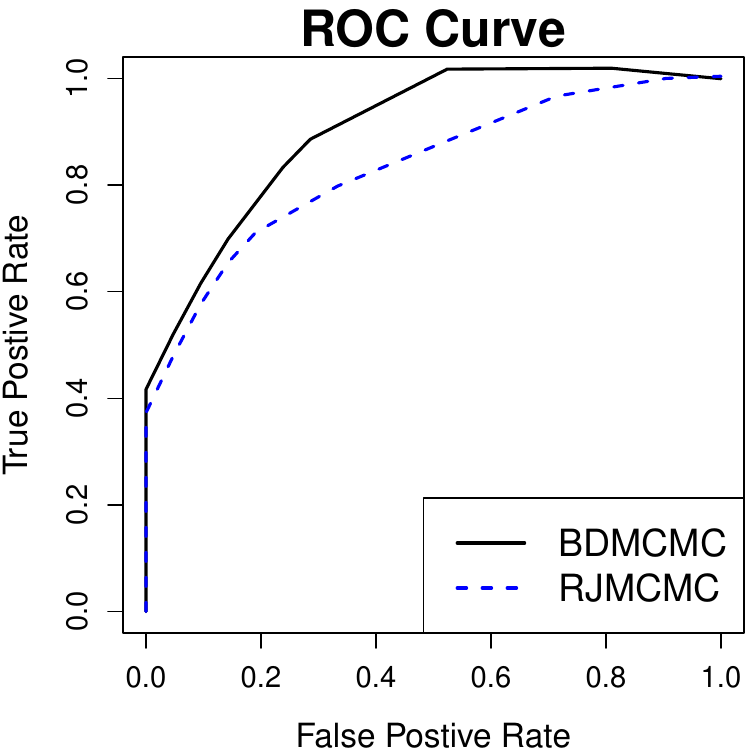}
\caption{ ROC plot to compare the performance of the BDMCMC and RJMCMC algorithms for a simulated toy example. }
\label{fig:plotroc}
\end{figure}

Users can compare the performance of these two algorithms by using the code
\begin{Sinput}
R> plotroc( data.sim, sample.bdmcmc, sample.rjmcmc, smooth = TRUE ) 
\end{Sinput}
which visualizes an ROC plot for both algorithms, BDMCMC and RJMCMC; see Figure \ref{fig:plotroc}.

We can also compare the performance of those algorithms by using the \code{compare} function as follows:
\begin{Sinput}
R> compare( data.sim, sample.bdmcmc, sample.rjmcmc, 
+           main = c( "True graph", "BDMCMC", "RJMCMC" ) )
\end{Sinput}
\begin{CodeOutput}
                True graph BDMCMC RJMCMC
true positive       7  5.000  5.000
true negative      21 20.000 19.000
false positive      0  1.000  2.000
false negative      0  2.000  2.000
F1-score            1  0.769  0.714
specificity         1  0.952  0.905
sensitivity         1  0.714  0.714
MCC                 1  0.704  0.619
\end{CodeOutput}
The results show that for this specific simulated example both algorithms have more or less the same performance; See \citet[Section 4]{mohammadi2015bayesianStructure}, \citet[Section 2.2.3]{mohammadi2016bayesian}. %  and the Supplementary material for a comprehensive simulation study.

\begin{figure} [!ht] % [!ht]
    \centering
    \includegraphics[width=1\textwidth]{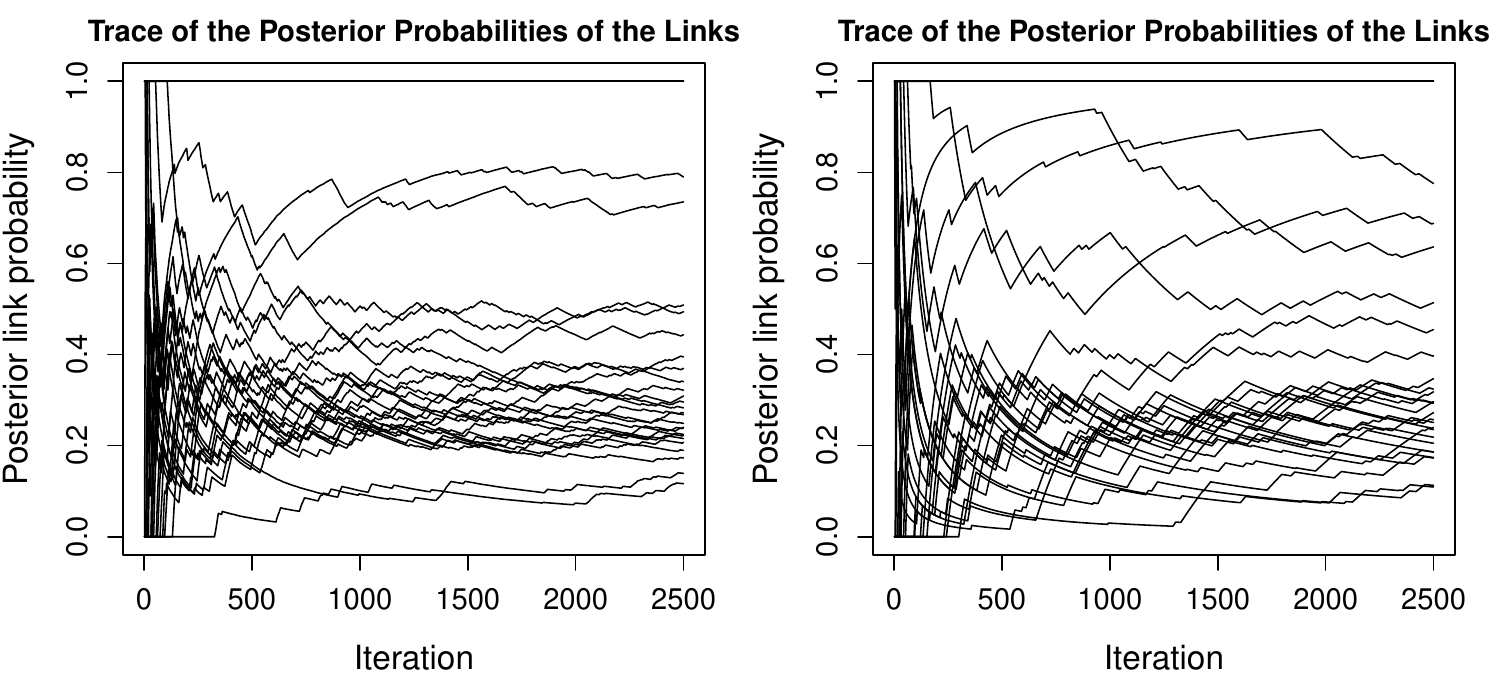}
\caption{
Plot for monitoring the convergence based on the trace of estimated posterior probability of all possible links for the BDMCMC algorithm (left) and the RJMCMC algorithm (right).
}
\label{fig:plotcoda}
\end{figure}

In this simulation example, we run both BDMCMC and RJMCMC algorithms for $5,000$ iterations, $2,500$ of them as burn-in. To check whether the number of iterations is enough and to monitoring the convergence of our both algorithm, we run
\begin{Sinput}
R> plotcoda( sample.bdmcmc )
R> plotcoda( sample.rjmcmc )
\end{Sinput}
The results in Figure \ref{fig:plotcoda} indicate that our BDMCMC algorithm converges faster with compare with RJMCMC algorithm. 

% - - - - - - - - - - - - - - - - - - - - - - - - - - - - - - - - - - - - - - - - - - - - - - - - - - - - - - - - - - - - - - - - - - - - - - - - - - - - - - - - - - - - |
\section{ Application to real datasets }
\label{sec:real data}

In this section we analyze two datasets from genetics and sociology, using the functions available in the \pkg{BDgraph} package. In Section \ref{subsec:survey data} we analyze a labor force survey dataset, involving mixed data. In Section \ref{subsec:gene data} we analyze human gene expression data, which do not follow the Gaussianity assumption. Both datasets are available in the \pkg{BDgraph} package.

% - - - - - - - - - - - - - - - - - - - - - - - - - - - - - - - - - - - - - - - - - - - - - - - - - - - - - - - - - - - - - - - - - - - - - - - - - - - - - - - - - - - - |
\subsection{ Application to labor force survey data }
\label{subsec:survey data}

\cite{hoff2007extending} analyzes the multivariate associations among income, education and family background, using data concerning $1002$ males in the U.S labor force. The dataset is available in the \pkg{BDgraph} package. 
% Users can call up the data via
\begin{Sinput}
R> data( "surveyData" )
R> head( surveyData, 5 )
\end{Sinput}
\begin{CodeOutput}
     income degree children pincome pdegree pchildren age
[1,]     NA      1        3       3       1         5  59
[2,]     11      0        3      NA       0         7  59
[3,]      8      1        1      NA       0         9  25
[4,]     25      3        2      NA       0         5  55
[5,]    100      3        2       4       3         2  56
\end{CodeOutput}    
Missing data are indicated by \code{NA}; in general, the rate of missing data is about $9$\%, with higher rates for the variables \texttt{income} and \texttt{pincome}. In this dataset we have seven observed variables as follows:
\begin{description} 
  \setlength{\itemsep}{0pt}
  \setlength{\parskip}{0pt}
  \setlength{\parsep}{0pt}
 \item \texttt{income}: An ordinal variable indicating respondent's income in 1000s of dollars.
 \item \texttt{degree}: An ordinal variable with five categories indicating respondent's highest educational degree.
 \item \texttt{children}: A count variable indicating number of children of respondent.
 \item \texttt{pincome}: An ordinal variable with five categories indicating financial status of respondent's parents.
 \item \texttt{pdegree}: An ordinal variable with five categories indicating highest educational degree of respondent's parents.
 \item \texttt{pchildren}: A count variable indicating number of children of respondent's parents.
 \item \texttt{age}: A count variable indicating respondent's age in years.
\end{description}
Since the variables are measured on various scales, the marginal distributions are heterogeneous, which makes the study of their joint distribution very challenging. However, we can apply to this dataset our Bayesian framework based on the Gaussian copula graphical models. Thus, we run the function \code{bdgraph} with option \code{method = "gcgm"}. For the prior distributions of the graph and precision matrix, as default of the function \code{bdgraph}, we place a uniform distribution as a uninformative prior on the graph and a G-Wishart $W_G(3,I_{7})$ on the precision matrix. We run our function for $10,000$ iterations with $7,000$ as burn-in.
\begin{Sinput}
R> sample.bdmcmc <- bdgraph( data = surveyData, method = "gcgm", 
+                            iter = 10000, burnin = 7000 )
R> summary( sample.bdmcmc )
\end{Sinput}
\begin{CodeOutput}
$selected_g
          income degree children pincome pdegree pchildren age
income         0      1        1       0       0         0   1
degree         0      0        1       0       1         1   0
children       0      0        0       0       1         1   1
pincome        0      0        0       0       1         0   0
pdegree        0      0        0       0       0         1   1
pchildren      0      0        0       0       0         0   0
age            0      0        0       0       0         0   0

$p_links
          income degree children pincome pdegree pchildren  age
income         0      1     1.00    0.37    0.06      0.05 1.00
degree         0      0     0.67    0.20    1.00      0.78 0.16
children       0      0     0.00    0.34    0.72      1.00 1.00
pincome        0      0     0.00    0.00    1.00      0.40 0.09
pdegree        0      0     0.00    0.00    0.00      0.92 0.99
pchildren      0      0     0.00    0.00    0.00      0.00 0.05
age            0      0     0.00    0.00    0.00      0.00 0.00

$K_hat
          income degree children pincome pdegree pchildren   age
income      1.33  -1.46    -0.54   -0.10    0.00      0.00 -0.33
degree     -1.46   7.63     0.46    0.08   -1.20      0.23 -0.04
children   -0.54   0.46     7.21    0.19    0.26     -0.40 -1.81
pincome    -0.10   0.08     0.19    6.92   -1.09      0.13  0.01
pdegree     0.00  -1.20     0.26   -1.09    1.36      0.20  0.22
pchildren   0.00   0.23    -0.40    0.13    0.20      1.17  0.00
age        -0.33  -0.04    -1.81    0.01    0.22      0.00  1.79
\end{CodeOutput}    
The results of the function \code{summary} are the adjacency matrix of the selected graph (\code{selected_g}), 
estimated posterior probabilities of all possible links (\code{p_links}) and estimated precision matrix (\code{K_hat}). 

\begin{figure} [!ht] % [!ht]
    \centering
    \includegraphics[width=0.65\textwidth]{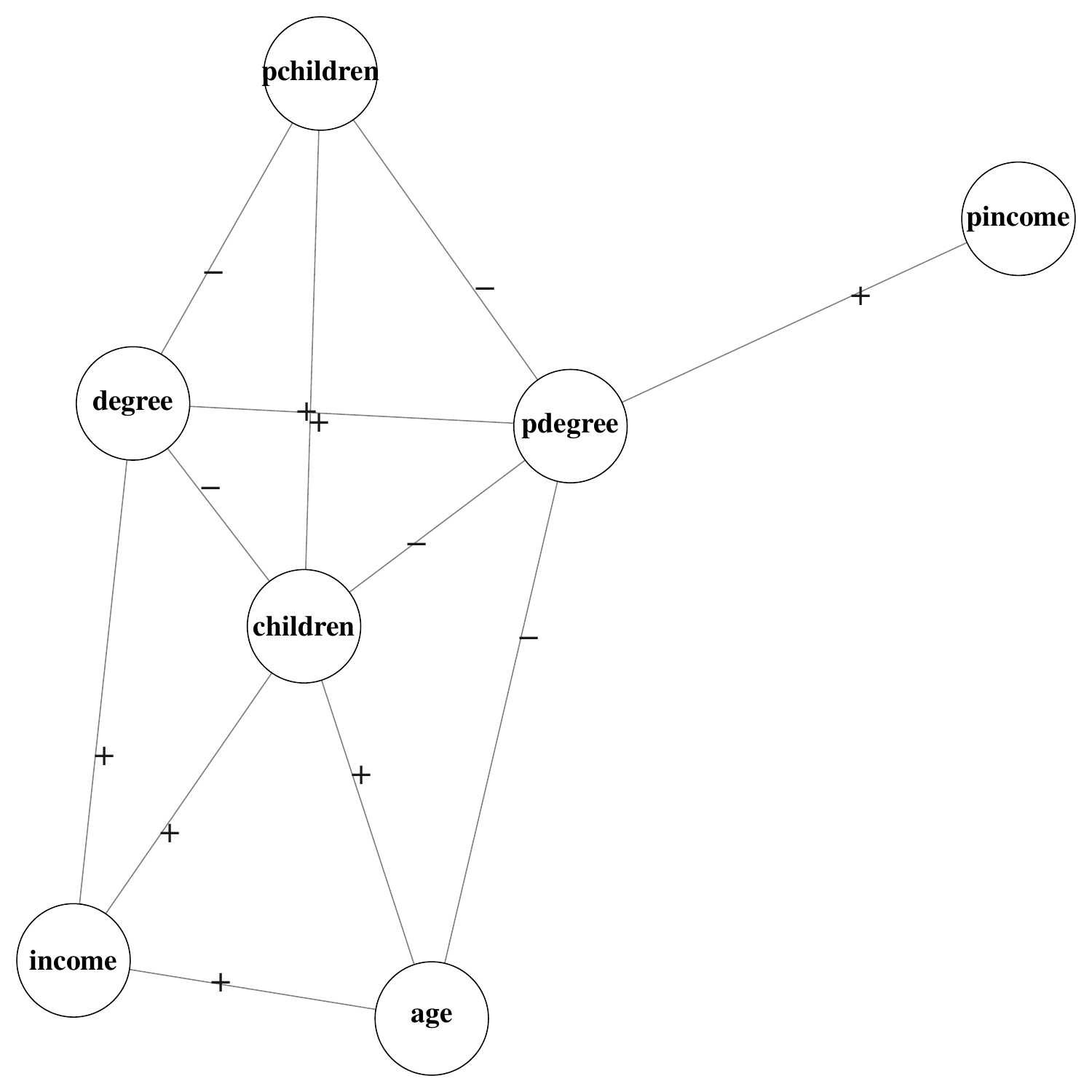}
\caption{
Inferred graph for the labor force survey data based on output from \code{bdgraph}. Sign ``+'' represents a positively correlated relationship between associated variables and ``-'' represents a negatively correlated relationship.
}
\label{fig:graph-surveydata}
\end{figure}
Figure~\ref{fig:graph-surveydata} presents the selected graph, a graph with links for which the estimated posterior probabilities are greater than $0.5$. Links in the graph are indicated by signs ``+''  and ``-'', which represent positively and negatively correlated relationships between associated variables, respectively. 

The results indicate that education, fertility and age have strong associations with income, since there are highly positively correlated relationships between income and those three variables, with posterior probability equal to one for all. It also shows that a respondent's education and fertility are negatively correlated, with posterior probability more than $0.67$. The respondent's education is certainly related to his parent's education, since there is a positively correlated relationship, with posterior probability equal to one. 
%Moreover, the results indicate that relationships between income, education and fertility hold across generations.
 
For this dataset, \cite{hoff2007extending} estimated the conditional independence between variables by inspecting whether the $95$\% credible intervals for the associated regression parameters, which do not contain zero. Our results are the same as that of Hoff except for two links. Our results indicate that there is a strong relationship between parents' education (\code{pdegree}) and fertility (\code{child}), a link which is not selected by Hoff. 

% - - - - - - - - - - - - - - - - - - - - - - - - - - - - - - - - - - - - - - - - - - - - - - - - - - - - - - - - - - - - - - - - - - - - - - - - - - - - - - - - - - - - |
\subsection{ Application to human gene expression }
\label{subsec:gene data}

Here, by using the functions that are available in the \pkg{BDgraph} package, we study the structure learning of the sparse graphs applied to the human gene expression data which were originally described by \cite{stranger2007population}. They collected data to measure gene expression in B-lymphocyte cells from Utah inhabitants with Northern and Western European ancestry. They considered $60$ individuals whose genotypes were available online at \url{ftp://ftp.sanger.ac.uk/pub/genevar}. Here the focus was on the $3,125$ Single Nucleotide Polymorphisms (SNPs) that were found in the 5' UTR (untranslated region) of mRNA (messenger RNA) with a minor allele frequency $\geq 0.1$. Since the UTR (untranslated region) of mRNA (messenger RNA) has previously been subject to investigation, it should play an important role in the regulation of gene expression. The raw data were background-corrected and  then quantile-normalized across replicates of a single individual and then median normalized across all individuals. 
Following \cite{bhadra2013joint}, of the $47,293$ total available probes, we consider the $100$ most variable probes that correspond to different Illumina TargetID transcripts. The data for these $100$ probes are available in our package. To see the data users can run the code
\begin{Sinput}
R> data( "geneExpression" )
R> dim( geneExpression )
\end{Sinput}
\begin{CodeOutput}
60 100
\end{CodeOutput}
The data consist of only $60$ observations ($n = 60$) across $100$ genes ($p = 100$). This dataset is an interesting case study for graph structure learning, as it has been used by \cite{bhadra2013joint, mohammadi2015bayesianStructure, gu2015local}.

\begin{figure}[!ht]
\centering
\includegraphics[width=0.8\textwidth]{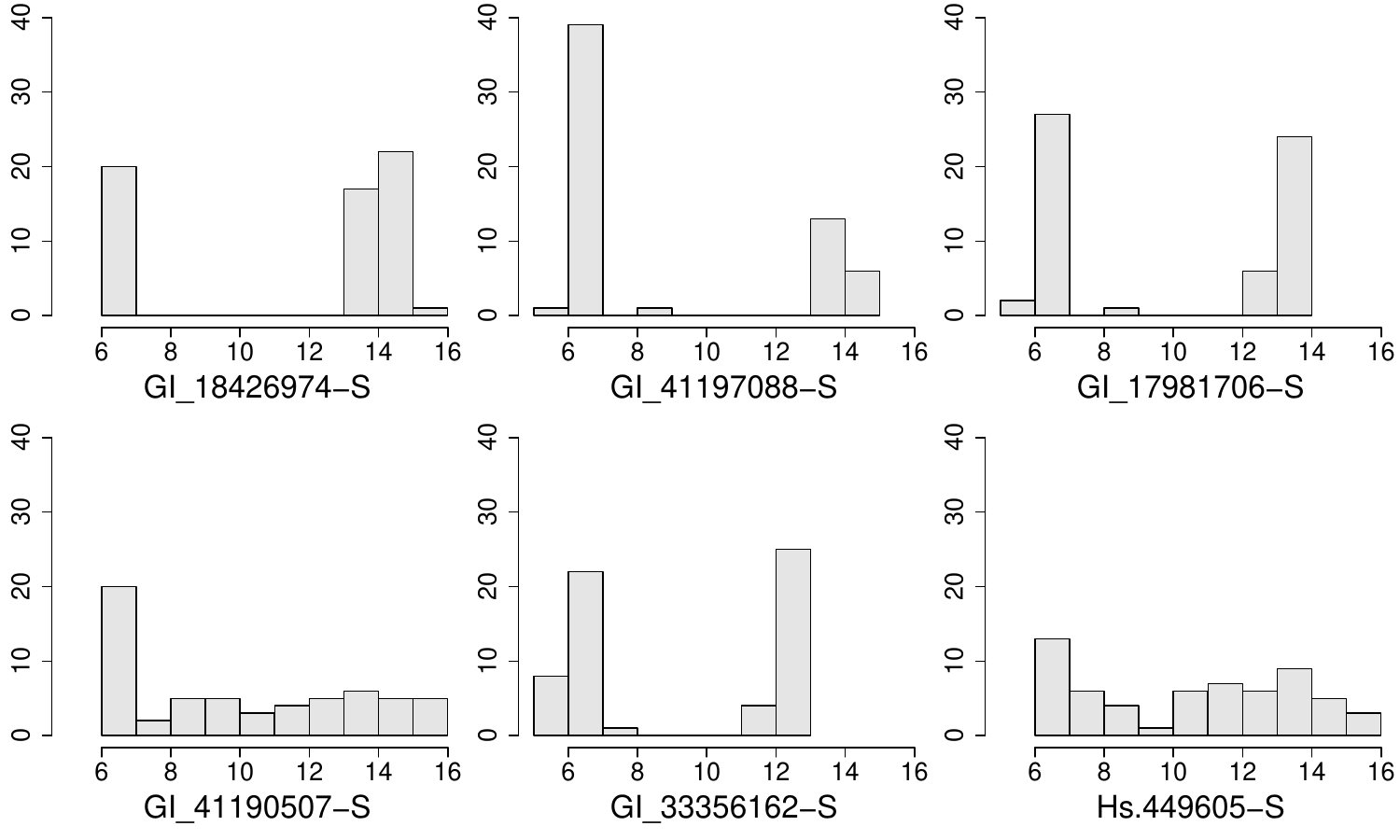}
\caption{ \label{fig:hist plot} 
Univariate histograms of first 6 genes in human gene dataset.}
\end{figure}

In this dataset, all the variables are continuous but not Gaussian, as can be seen in Figure \ref{fig:hist plot}. Thus, we apply Gaussian copula graphical models, using the function \code{bdgraph} with option \code{method = "gcgm"}. For the prior distributions of the graph we use a Bernoulli prior on each link inclusion \eqref{graph prior}, encourage sparsity by considering $\theta = 0.1$, using the function \code{bdgraph} with option \code{g.prior = 0.1}. For the prior distributions of the precision matrix, as default of the function \code{bdgraph}, we place the G-Wishart $W_G(3,I_{100})$ on the precision matrix. 

\begin{figure}[!ht]
\centering
\includegraphics[width=0.7\textwidth]{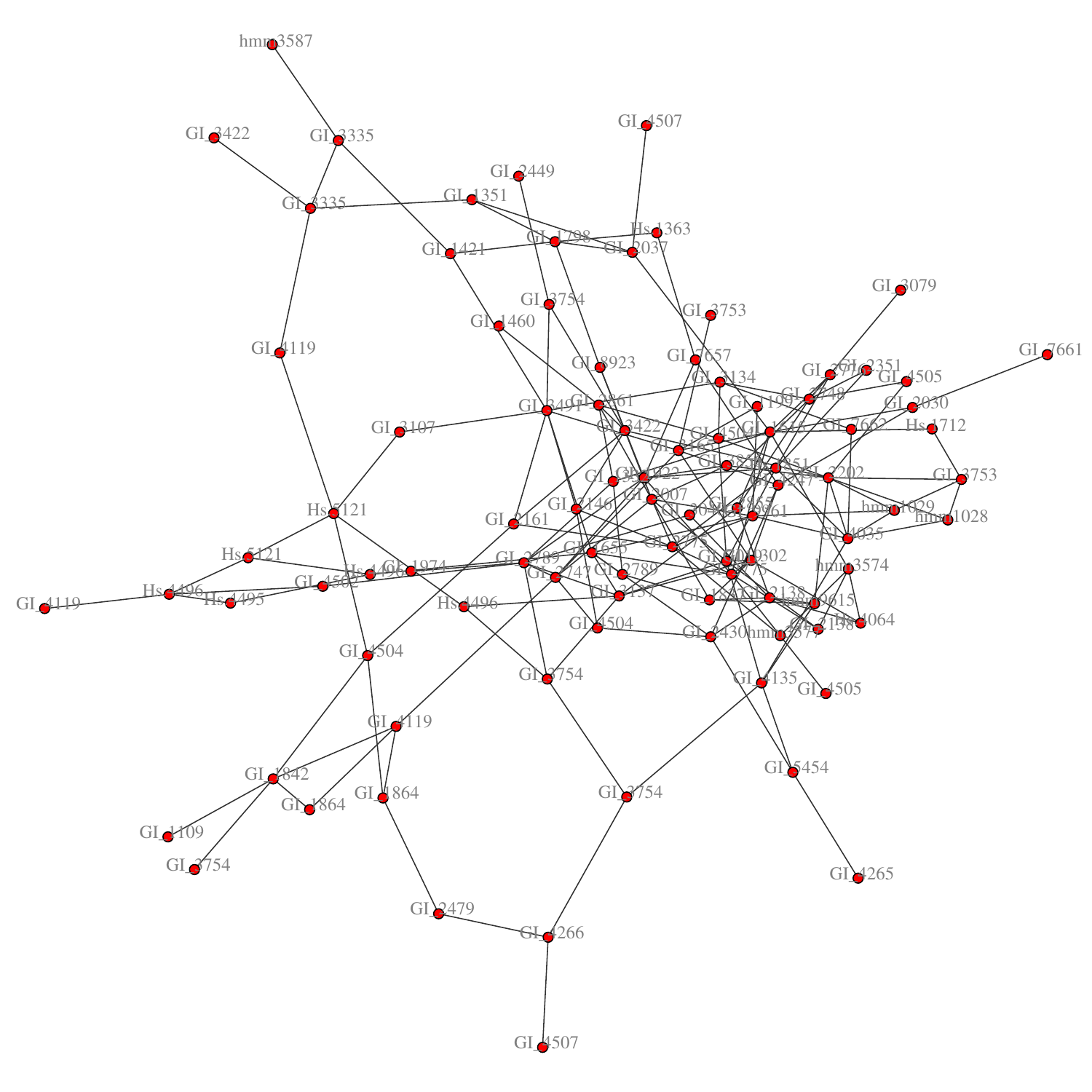}
\caption{ \label{fig:graph-gene data} 
The inferred graph for human gene expression data using Gaussian copula graphical models. This graph consists of $176$ links with estimated posterior probabilities greater than $0.5$.}
\end{figure}

We run our function for $10,000$ iterations with $7,000$ as burn-in as follows: 
\begin{Sinput}
R> sample.bdmcmc <- bdgraph( data = geneExpression, method = "gcgm", 
+                            g.prior = 0.1, iter = 10000, burnin = 7000 )       
\end{Sinput}
This took less than $3$ minutes. We use the following code to visualize the graph with estimated posterior probabilities greater than $0.5$. 
\begin{Sinput}
R> select( sample.bdmcmc, cut = 0.5, vis = TRUE )     
\end{Sinput}
By using option \code{vis = TRUE}, the function plots the selected graph. Figure \ref{fig:graph-gene data} visualizes the selected graph which consists of $176$ links with estimated posterior probabilities (\ref{posterior-link}) greater than $0.5$.
\begin{figure}[!ht]
\centering
\includegraphics[width=0.7\textwidth]{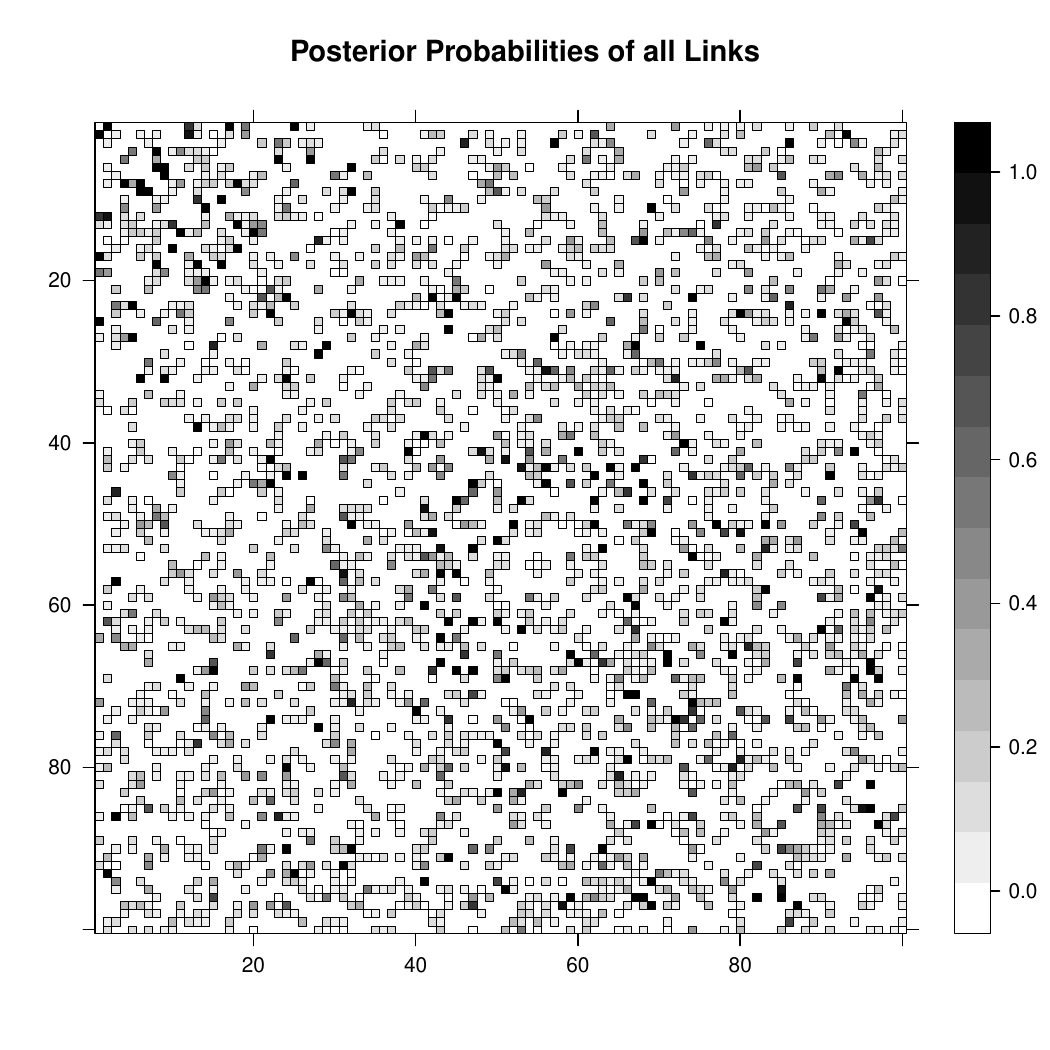}
\caption{ \label{fig:phat-gene data} 
Image visualization of the estimated posterior probabilities of all possible links in the graph on human gene expression data.}
\end{figure}

The function \code{plinks} reports the estimated posterior probabilities of all possible links in the graph. For our data the output of this function is  a $100 \times 100$ matrix. Figure \ref{fig:phat-gene data} reports the visualization of that matrix.

Most of the links in our selected graph conform to results in previous studies. For instance, \cite{bhadra2013joint} found $54$ significant interactions between genes, most of which are covered by our method. In addition, our approach indicates additional gene interactions with high posterior probabilities that are not found in previous studies; this result may complement the analysis of human gene interaction networks.

% - - - - - - - - - - - - - - - - - - - - - - - - - - - - - - - - - - - - - - - - - - - - - - - - - - - - - - - - - - - - - - - - - - - - - - - - - - - - - - - - - - - - |
\section{Conclusion}
\label{conclusion}

We have presented the \pkg{BDgraph} package which was designed for Bayesian structure learning in general -- decomposable and non-decomposable -- undirected graphical models. The package implements recent improvements in computation, sampling and inference of Gaussian graphical models \citep{mohammadi2015bayesianStructure, dobra2011bayesian} for Gaussian data and Gaussian copula graphical models \citep{mohammadi2016bayesian, dobra2011copula} for non-Gaussian, discrete and mixed data.

We are committed to maintaining and developing the \pkg{BDgraph} package in the future. Future versions of the package will contain more options for prior distributions of the graph and precision matrix. One possible extension of our package, would be to deal with outliers, by using robust Bayesian graphical modelling using Dirichlet \textbf{t}-Distributions \citep{finegold2014robust, mohammadi2014contribution}. An implementation of this method would be desirable in actual applications.
%Another insteasting extention of our package would be to apply in Bayesian regression tree, sence the tree is apacial case of graph \citep{

% - - - - - - - - - - - - - - - - - - - - - - - - - - - - - - - - - - - - - - - - - - - - - - - - - - - - - - - - - - - - - - - - - - - - - - - - - - - - - - - - - - - - |
\section*{Acknowledgments}

The authors are grateful to the associated editor and reviewers for helpful criticism of the original of both the manuscript and the \proglang{R} package. We would like to thank Sven Baars for parallel implementation in \proglang{C++}. We also would like to thank Sourabh Kotnala for implementing the package in \proglang{C++}.

% - - - - - - - - - - - - - - - - - - - - - - - - - - - - - - - - - - - - - - - - - - - - - - - - - - - - - - - - - - - - - - - - - - - - - - - - - - - - - - - - - - - - |
\section*{Appendix: Dealing with memory usage restriction}
\label{appendix}

The memory usage restriction is one of the challenges of Bayesian inference for maximum a posterior probability (MAP) estimation and monitoring the convergence check, especially for high-dimensional problems. For example, to compute MAP estimation in our \pkg{BDgraph} package, we must document the adjacency matrices of all the visited graphs by our MCMC sampling algorithms, which causes memory usage restriction in \proglang{R}. Indeed, the function \code{bdgraph} in our package in case \code{save = TRUE} is documented to return all of the adjacency matrices for all iterations after burn-in. For instance, for the case
\begin{Sinput}
R> iter   <- 10000
R> burnin <- 7000
R> p      <- 100
R> graph  <- matrix( 1, p, p ) 
R> print( ( iter - burnin ) * object.size( graph ), units = "auto" ) 
\end{Sinput}
\begin{CodeOutput}
3.7 Gb
\end{CodeOutput}
A naive way is to save all the matrices, which leads to memory usage restriction, as it costs $3.7$ gigabytes of memory. To cope with this problem, instead of saving all adjacency matrices we simply transfer the upper triangular part of the adjacency matrix to one single character; see codes below:
\begin{Sinput}
R> string_graph <- paste( graph[ upper.tri( graph ) ], collapse = '' )
R> print( ( iter - burnin ) * object.size( string_graph ), units = "auto" ) 
\end{Sinput}
\begin{CodeOutput}
241.1 Mb
\end{CodeOutput}
In this efficient way we need only $241.1$ megabytes instead of $3.7$ gigabytes of memory.

% - - - - - - - - - - - - - - - - - - - - - - - - - - - - - - - - - - - - - - - - - - - - - - - - - - - - - - - - - - - - - - - - - - - - - - - - - - - - - - - - - - - - |
\bibliographystyle{Chicago}
\bibliography{ref_BDgraph}
% - - - - - - - - - - - - - - - - - - - - - - - - - - - - - - - - - - - - - - - - - - - - - - - - - - - - - - - - - - - - - - - - - - - - - - - - - - - - - - - - - - - - |

% - - - - - - - - - - - - - - - - - - - - - - - - - - - - - - - - - - - - - - - - - - - - - - - - - - - - - - - - - - - - - - - - - - - - - - - - - - - - - - - - - - - - |
\end{document}